\documentclass[runningheads]{llncs}
\usepackage[T1]{fontenc}
\usepackage{graphicx}
\usepackage{amsmath}
\usepackage[ruled,vlined]{algorithm2e}
\usepackage[numbers]{natbib}
\usepackage[utf8]{inputenc} % allow utf-8 input
\usepackage[T1]{fontenc}    % use 8-bit T1 fonts
\usepackage{hyperref}       % hyperlinks
\usepackage{url}            % simple URL typesetting
\usepackage{booktabs}       % professional-quality tables
\usepackage{amsfonts}       % blackboard math symbols
\usepackage{nicefrac}       % compact symbols for 1/2, etc.
\usepackage{microtype}      % microtypography
\usepackage{lipsum}
\usepackage{fancyhdr}       % header
\usepackage{graphicx}       % graphics
\graphicspath{{media/}}     % organize your images and other figures under media/ folder
\usepackage{amsmath}
\usepackage{amssymb}
\usepackage{breqn}
\usepackage{paralist}
\usepackage[ruled,vlined]{algorithm2e}
\usepackage{xcolor}
\usepackage{subfig}
\usepackage{comment}
\usepackage{appendix}
\usepackage{multirow}
\usepackage{siunitx}

\begin{document}
\title{Domain penalisation for improved Out-of-Distribution Generalisation}
\author{Shuvam Jena\inst{1}\orcidID{0009-0006-8098-3052} \and
        Sushmetha Sumathi Rajendran\inst{1}\orcidID{0009-0007-1185-8255} \and
        Karthik Seemakurthy\inst{2}\orcidID{0000-0003-2954-6149} \and
        Sasithradevi A \inst{1}\orcidID{0000-0001-5198-6648} \and
        Vijayalakshmi M \inst{1}\orcidID{0000-0002-3751-7041} \and
        Prakash Poornachari \inst{3}\orcidID{0000-0002-7534-6925}
        }

\authorrunning{Shuvam et al.}
\institute{Vellore Institute of Technology, Chennai, India.  \and
Hydronium Energies, London, United Kingdom. \and
Anna University, Chennai, India.}
\maketitle              % typeset the header of the contribution
\begin{abstract}
In the field of object detection, domain generalisation (DG) aims to ensure robust performance across diverse and unseen target domains by learning the robust domain-invariant features corresponding to the objects of interest across multiple source domains. While there are many approaches established for performing DG for the task of classification, there has been a very little focus on object detection. In this paper, we propose a domain penalisation (DP) framework for the task of object detection, where the data is assumed to be sampled from multiple source domains and tested on completely unseen test domains. We assign penalisation weights to each domain, with the values updated based on the detection network's performance on the respective source domains. By prioritising the domains that needs more attention, our approach effectively balances the training process. We evaluate our solution on the GWHD 2021 dataset, a component of the WiLDS benchmark and we compare against ERM and GroupDRO as these are primarily loss function based. Our extensive experimental results reveals that the proposed approach improves the accuracy by 0.3~$\%$ and 0.5~$\%$ on validation and test out-of-distribution (OOD) sets, respectively for FasterRCNN. We also compare the performance of our approach on FCOS detector and show that our approach improves the baseline OOD performance over the existing approaches by 1.3~$\%$ and 1.4~$\%$ on validation and test sets, respectively. This study underscores the potential of performance-based domain penalisation in enhancing the generalisation ability of object detection models across diverse environments.

\end{abstract}
\section{Introduction}
\label{sec:introduction}
\begin{figure}
   \centering
    \begin{tabular}{cccc}
     \includegraphics[scale=0.075]{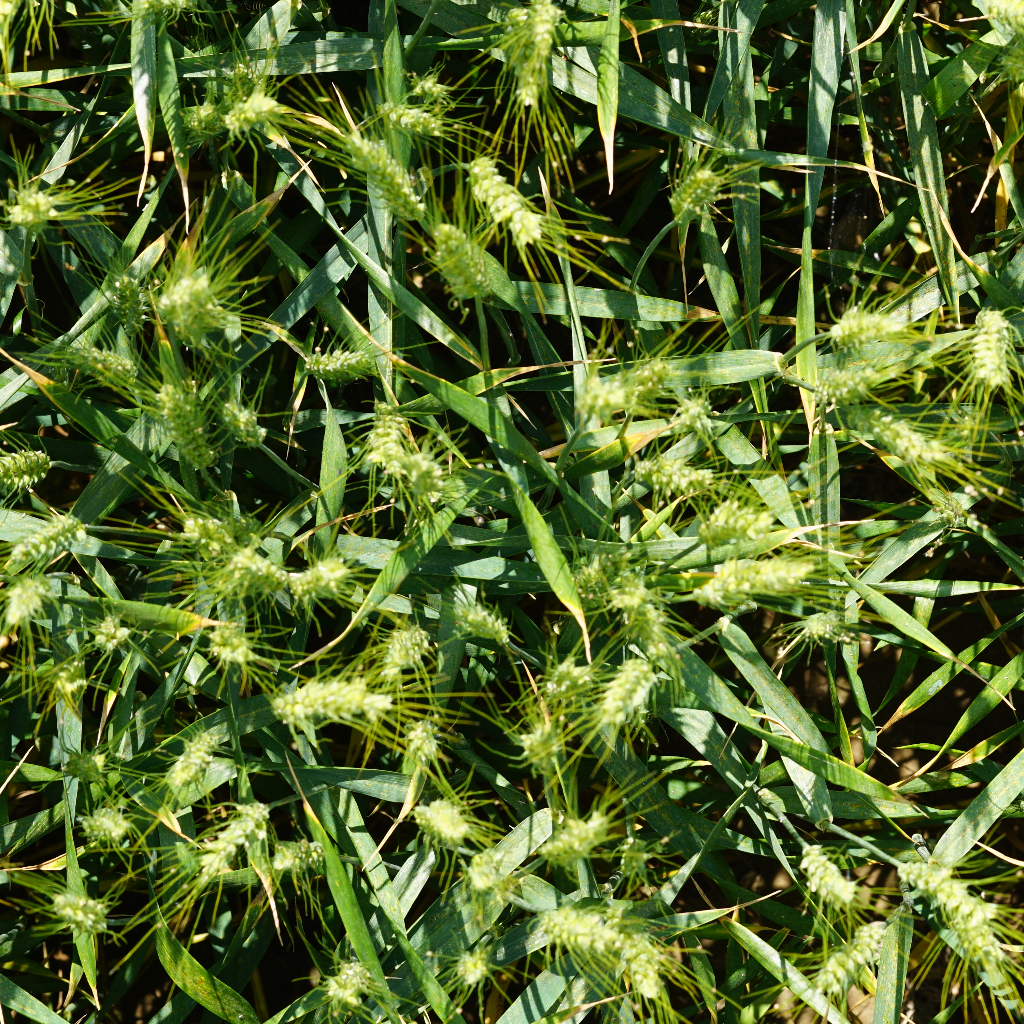} &
     \includegraphics[scale=0.075]{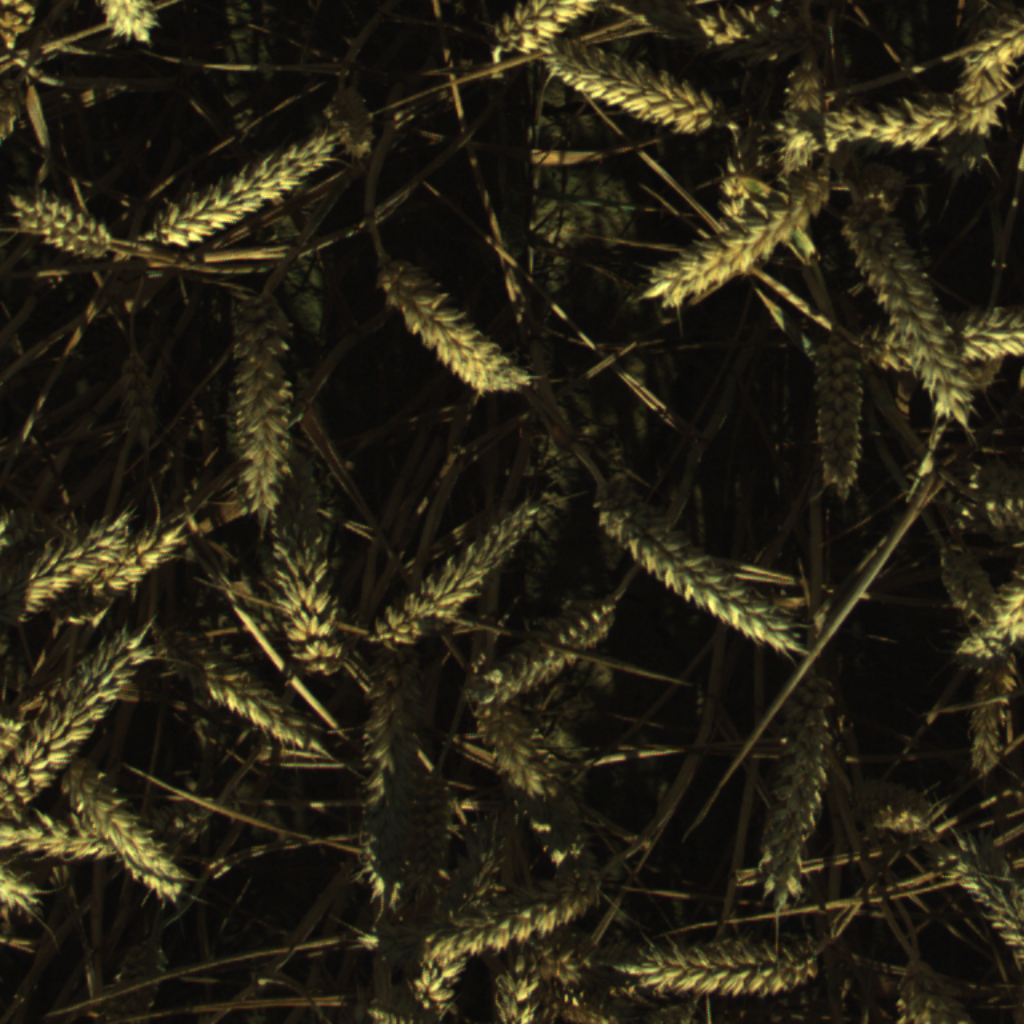} &
     \includegraphics[scale=0.075]{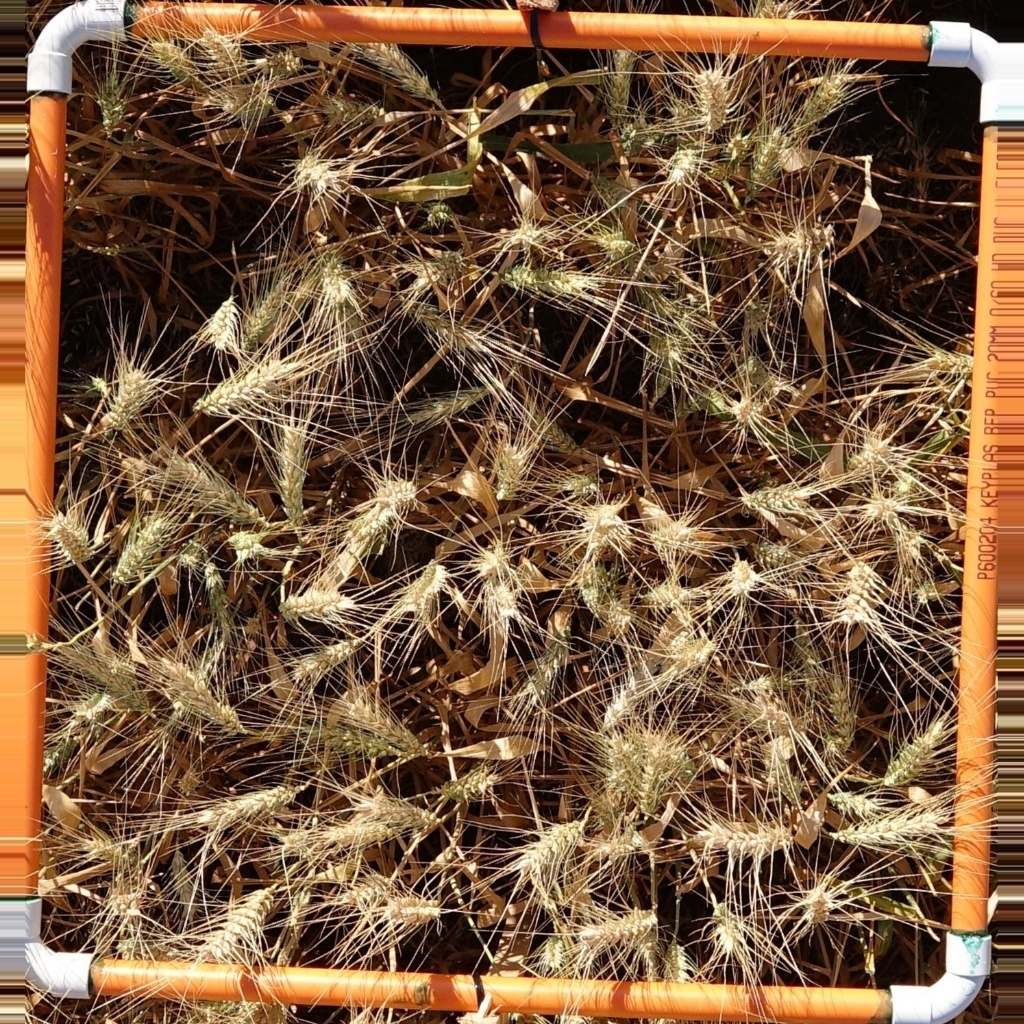} &
     \includegraphics[scale=0.075]{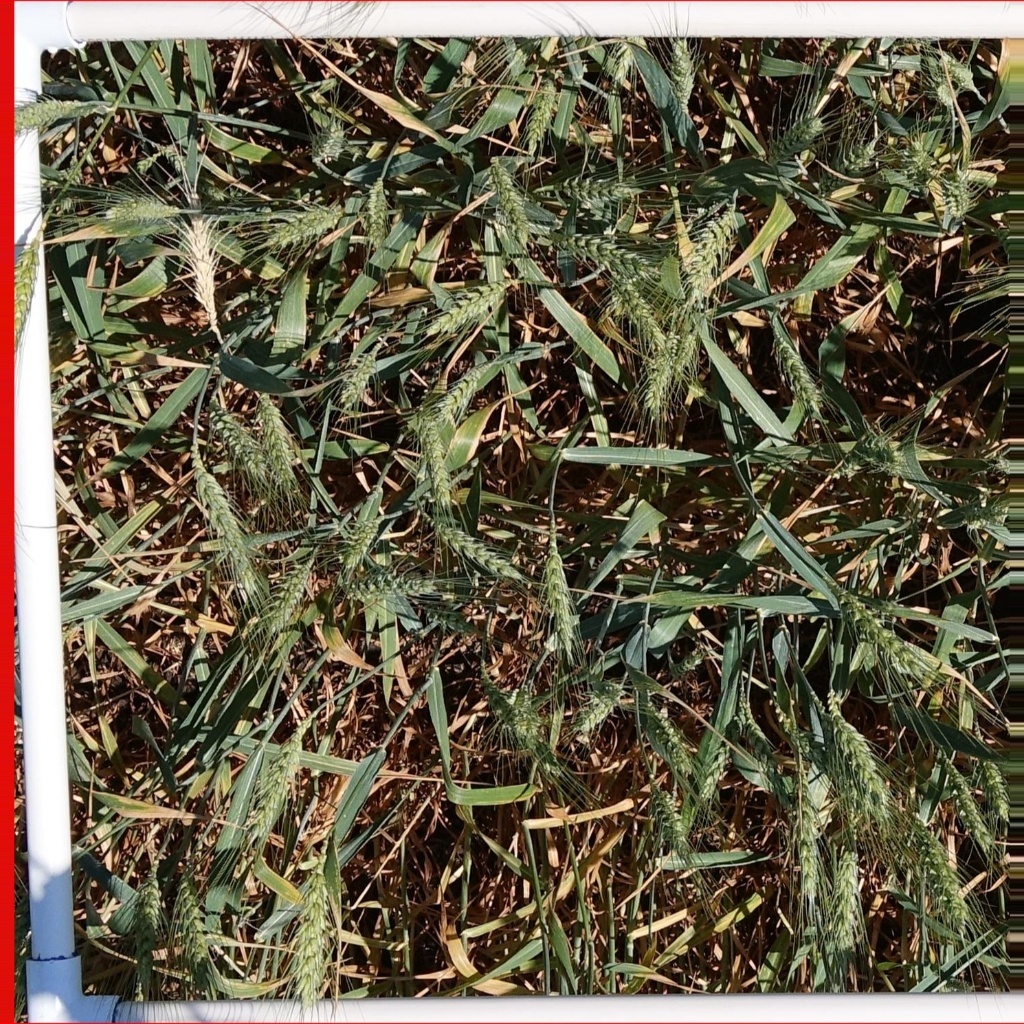} \\
     Rres_1 & Arvalis_12 & UQ_7 & UQ_10
    \end{tabular}
    \caption{Sample images from GWHD 2021 dataset. Training domains: Rres_1 and Arvalis_12. Test domains: UQ_7 and UQ_10.}
    \label{fig:intro_ex}
\end{figure}

In computer vision object detection is a cornerstone with crucial applications in many industries. Its practical value has driven a surge in research and development in recent years. Deep learning based frameworks like Faster R-CNN~\cite{faster2015towards}, YOLO~\cite{redmon2016you}, DETR~\cite{carion2020end}, and FCOS~\cite{lin2017focal} have exceeded conventional approaches that relied on basic classifiers and manual feature extraction. These advanced approaches achieve superior performance in complex settings, showcasing cutting-edge capabilities. Even though deep learning based approaches were successful in terms of accuracy, their real time performance is primarily impacted by the domain shift which is manifested in several different ways. In medical applications, it is common practice to train a model on data from one hospital and then test it using data from another hospital ~\cite{zech2018variable}. In wildlife monitoring, researchers may aim to develop a model for animal recognition using images from a particular set of camera traps and subsequently deploy it to recognise wildlife from images captured from different camera traps~\cite{beery2018recognition}. In this work, we use the Global Wheat Head Detection (GWHD) 2021 dataset, where the model is trained on images captured in one set of farms and is deployed to detect wheat heads in a completely different set of farms~\cite{david2021global}. While there exists many object detection datasets, GWHD 2021 is the only multi-source domain dataset which has been benchmarked in WilDS. The domain shift for wheat heads is primarily due to differences in growth stages, cameras used, locations where the wheat is grown, and wheat density. Fig.~\ref{fig:intro_ex} shows sample images from the training and test sets of the GWHD 2021 dataset. It can be seen that there is significant diversity in the wheat heads. In the proposed approach, our aim is to reduce the domain gap in the GWHD 2021 and thereby enhancing the wheat head detection accuracy in the test environments where the model is finally deployed.

Majority of the existing deep learning models assume that the training and test data are sampled from a similar distribution. However, in many practical scenarios as seen in Fig.~\ref{fig:intro_ex}, there exists a significant domain gap in the training and test data distributions and this deteriorates the deployed model performance. In the recent past, there has been a significant focus towards bridging the domain gap between the  training and test data distributions~\cite{khoee2024domain, niu2024survey, nguyen2024tackling, karunanayake2024out, zhou2022domain, wang2022generalizing, liu2021towards}. While there exists many different ways to address the problem of domain shift as mentioned in these surveys, the proposed approach in this paper proposes to train the object detection networks to learn domain-invariant features for multi-source and multi-target domains scenario. Here we minimise the sum of weighted per-domain average losses and show that our approach can perform better over the techniques (ERM and GroupDRO) used within the WiLDS benchmark~\cite{koh2021wilds}. The weights assigned to each of the per-domain losses are dynamically computed based on the perfomance of the object detector on the respective source domains. The key contributions of the proposed approach are:
\begin{itemize}
    \item We proposed a domain penalisation framework where we assign weights for the average loss computed for each domain in object detection scenario. 
    \item We show that our framework is detector agnostic by demonstrating the improved performance on the two standard detectors FasterRCNN and FCOS.  
    \item We compare and show that our approach performs better over the standard loss function based techniques used in WiLDS benchmark (ERM, GroupDRO) for the object detection scenario. 
\end{itemize}

\section{Related works}
\label{sec:related_works}

In the recent past, there is a significant focus towards improving the generalisation ability of the deep neural networks. There are number of surveys, which summarise the state-of-the-art DG approaches~\cite{khoee2024domain, niu2024survey, nguyen2024tackling, karunanayake2024out, zhou2022domain, wang2022generalizing, liu2021towards}. While some of these surveys are focused towards a certain specific applications and few others attempt to give generic picture of various DG approaches. In this section, we will describe the related works of the approaches used for improving the generalisation ability of the object detectors in specific.   

One of the most earlier approaches which attempted to bridge the domain shift is transfer learning (TL)~\cite{pan2009survey}, which leverages the high transferability of pre-trained deep features, particularly from the lower layers of neural networks. This approach involves fine-tuning architectures that have been pre-trained on extensive datasets, such as ImageNet~\cite{deng2009imagenet} or COCO~\cite{lin2014microsoft}, for a different yet related task. Unlike DG, TL necessitates annotated data from the target domain, but it significantly reduces both the time and the number of annotations required compared to training a model from scratch. The performance of TL based approaches degrades with the increase in domain shift. In contrast to TL, which require fully annotated target domain data, self-supervised learning (SSL)~\cite{jing2020self} leverages unlabeled data to minimize the shift in feature space. This can be achieved sequentially through a pre-training step followed by classical supervised TL~\cite{noroozi2016unsupervised, hindel2023inod} or in a semi-supervised multi-task learning setup, where unlabeled data is used alongside a smaller quantity of labeled data. The secondary task in SSL, which relies on unlabeled data, has a regularizing effect on the learned features and has also been applied for DG~\cite{ilteralp2021deep, ullah2023ssmd, li2022learning}. Domain randomization~\cite{tobin2017domain, yue2019domain} is another complementary approach to DG, which involves generating synthetic variations of the input data to obtain more generalizable features. Domain adaptation (DA)~\cite{saenko2010adapting, chen2018domain} is closely related to DG, as both aim to address the domain shift encountered in new environments without altering the label space. However, unlike DG, where no information about the target data distribution is available, DA utilizes sparsely labeled or unlabeled target domain data to align the feature spaces corresponding to source and target domains.

While there exists many approaches to bridge the domain shift between the training and test distributions, there exists a very few approaches in the context of object detection~\cite{liu2020towards, lin2021domain, seemakurthy2023domain_fcos, seemakurthy2023domain_frcnn, koh2021wilds}. However, majority of these techniques are primarily focused towards the architectural modifications of the standard object detectors to achieve DG. Unlike these approaches, we propose a weighted loss function to achieve DG. 
Some of the popular approaches in this category are used in the WilDS benchmark~\cite{koh2021wilds}. Empirical Risk Minimisation (ERM) trains the model by minimising the average training loss~\cite{koh2021wilds} and do not consider any form of additional meta-data (domain) information from the dataset. There are few other approaches which adds penality to the ERM based objective to encourage the deep networks to learn domain invariances~\cite{sun2016deep, arjovsky2019invariant}. Correlation Alignment~(CORAL)~\cite{sun2016deep} deals with minimisation of means and covariances of various feature distributions across the domains. CORAL is similar to many other approaches which encourage the network to equalise the feature distributions across multiple source domains~\cite{tzeng2014deep, long2015learning, ganin2016domain, li2018deep}.  Invariant Risk Minimisation (IRM) penalises the feature distributions which got different optimal linear classifiers for each domain~\cite{arjovsky2019invariant}. Group Distributionally Robust Optimisation~(GroupDRO) penalises the worst performing domain during training~\cite{hu2018does, sagawa2019distributionally}. The proposed approach in this paper is primarily inspired from GroupDRO. However, instead of minimising the worst group accuracy alone, we propose a new loss function which computes the weighted sum of losses correspond to each of the source domains. The weights for each loss component is computed based on the performance of the detector on each of the source domains. Through our experimental results, we prove that our strategy can perform better over the techniques used for demonstrating the generalisation abilities of the object detectors in the WiLDS benchmark.     

\section{Proposed Approach}
\label{sec:proposed_approach}
This section describes the details of the mathematical framework of our proposed approach. We deal with the problem of domain penalisation with an aim to improve out-of-distribution (OOD) performance of the deep learning networks.  Here we assume that the training data is sampled from multiple source domains which is often encountered in many computer vision applications. For example: multiple source domains can either correspond to the data collected from multiple weather conditions in autonomous driving scenario or the data collected from multiple different countries~\cite{koh2021wilds}. In this paper, we deal with improving the OOD performance for the multi-source domain object detection dataset GWHD 2021 from the WiLDS benchmark~\cite{koh2021wilds}. 

Let $D_i$ denote the $i^{th}$ domain while $\mathbf{I}_j^{D_i}$ be the $j^{th}$ image sampled from the $i^{th}$ domain. We assume our training dataset comprises a total of $N_s$ source domains. The proposed approach is primarily inspired from the team games whose winning depends on the performance of all the individual team members. Here each of the source domains can be considered as the individual players and we show that the OOD performance of the object detection networks can be improved by training the deep networks in such a way that its performance improves on all of the source domains simultaneously as the training progresses. We achieve this through appropriate penalisation weights to each of the source domains based on how the object detector performs on each of these source domains.  

Let $w_i$ be the weight assigned to the $i^{th}$ training domain and these weights are computed based on the object detector performance on each of the source domains. There are two main steps in the proposed approach. The loss at each training step for each batch is computed as the weighted sum of per-image loss as follows:
\begin{equation}
    L_B = \sum_{i=1}^{N_B} w_i * L_{ij}
    \label{eq:training_step}
\end{equation}
where $N_B$ indicates the batchsize, $L_{ij}$ denotes the loss corresponding to the $\mathbf{I}_j^{D_i}$ and $L_B$ denotes the loss corresponding to the batch B. The loss $L_B$ will be used for training the network and we also keep track of the average domain loss $L_{D_i}$ for each of the source domain $D_i$ which is computed as follows:  
\begin{equation}
L_{D_i} = \frac{1}{M_i} \sum_{j=1}^{M_i} L_{ij}
\label{eq:average_domain_loss}
\end{equation}
where $M_i$ correspond to the number of images in $i^{th}$ domain. The weights in Eq.~\ref{eq:training_step} are initialised by unity in the first epoch and is computed by using softmax as follows in the subsequent epochs:
\begin{equation}
    w_i = \frac{\exp(L_{D_i})}{\sum_{i=1}^{M_i} \exp(L_{D_i})}
    \label{eq:softmax}
\end{equation}
Note that we calculate the weight $w_i$ by using the average domain loss.  
The higher the average loss corresponding to a specific domain, the more it would be penalised in the subsequent epochs and this implies that the model is expected to focus more on the domains which have higher average training loss. The proposed approach is summarised in Algorithm~\ref{alg:Alg1}. 
\begin{algorithm}
\SetAlgoLined
\textbf{Input:} M source domain training datasets, $w_i$ = 1, $\forall$i \\ 
\textbf{Output:} Trained object detection model \\
\For{each epoch}{
 \For{each batch}{
  Randomly sample a batch of data from the training dataset\\
  Compute the loss using the Eq.~\ref{eq:training_step}\\
  Backpropogate the loss to train the model \\
  }
  Compute the domain penalisation weights by using the Eq.~\ref{eq:average_domain_loss}\\
  Normalise the weights by using the Eq.~\ref{eq:softmax}
}
\caption{Domain penalisation for improved OOD Generalisation.}
\label{alg:Alg1}
\end{algorithm}

\section{Experimental Results}
\label{sec:experimental_results}
In this section, we describe the details of our experimental results and demonstrate the improved generalisation performance of the proposed framework on the GWHD 2021 dataset~\cite{david2021global}.

\textbf{Dataset details:} GWHD 2021 dataset stands as a promising  resource in the field of deep learning-based wheat head detection. This dataset is aimed towards improving the efficiency of the tasks like plant phenotyping and precision agriculture through automation. It stands alone as the first comprehensive dataset designed specifically for multi-source domain generalisation. This unique quality empowers researchers to develop robust models that excel in real-world scenarios with diverse environments. GWHD 2021 comprises of over 6000 high-resolution images (1024 $\times$ 1024 pixels) meticulously annotated with bounding boxes for more than 300K unique wheat heads. Captured across a wide range of 11 countries and 47 measurement sessions, these images encompass a vast spectrum of real-world conditions. From varying weather patterns and times of day to diverse geographic locations and sensor types, the dataset offers unparalleled diversity. To effectively measure the generalisation performance of the proposed approach, we use the following splits as recommended in the WilDS benchmark~\cite{koh2021wilds}. 
\begin{itemize}
    \item \textbf{Official Train:} The dataset comprises images from 18 collection sessions conducted in various locations across Europe, which includes 13 sessions from France, Norway(2 sessions), 1 session from Switzerland, the United Kingdom, and Belgium. In total, there are 2,943 images featuring 131,864 wheat heads.
    \item \textbf{Official Validation:} This split includes images from 7 different collection sessions in Asia (4 in Japan and 3 in China) and 1 session in Africa (Sudan). Altogether, there are 1,424 images featuring a total of 44,873 wheat heads.
    \item  \textbf{Official Test:} This split consists of images from 11 collection sessions in Australia and 10 other sessions in North America, which include the USA (6 sessions), Mexico (3 sessions), and Canada. In total, there are 1,434 images featuring 66,905 wheat heads.
    \item \textbf{Mixed Train:} This split consists of images from 18 collection sessions in Europe, which include France (13 sessions), Belgium (1 session), Norway (2 sessions), Switzerland (1 session), and the UK (1 session), 1 session in Sudan, Africa, 11 sessions in Australia, 7 sessions in Asia, which include Japan (4 sessions) and China (3 sessions), and 2 sessions in Mexico. In total, there are 2,943 images featuring 133,334 wheat heads across 39 sessions.
    \item \textbf{Mixed Test:} This split consists of images from 11 acquisition sessions in Australia, 1 session in Africa (Sudan), 7 sessions in Asia, which include (3 from china and 4 from japan) and 2 sessions from South America (both from Mexico). In total, there are 720 images featuring 33,301 wheat heads across 21 sessions.
\end{itemize}

\textbf{Training details:} The proposed framework is detector network agnostic and we prove that our approach works well on the standard two-stage~(FasterRCNN) and single-stage network~(FCOS) models. We have used the models downloaded from the torchvision library. Even though we prove the improved OOD generalisation using our proposed approach on FasterRCNN and FCOS architectures, the same approach can be extended to any other object detection network architecture. We initialise both the networks with their respective COCO pretrained weights and use PyTorch framework for training the networks. In order to do fair comparison against the approaches used in the WiLDS benchmark, we use the learning rate of 0.0001, a batchsize of 4 and fixed the total number of 12 epochs. Optimization was carried out using the Adam optimizer. 
Every experiment was performed with three random seeds and the standard deviation has been reported along with the accuracy. We initially replicate the evaluations reported in the WiLDS benchmark and then report the performance of the proposed approach in all the experimental settings as reported in ~\cite{koh2021wilds}.

\textbf{Performance Metrics:} Following the OOD performance analysis in the WiLDS benchmark~\cite{koh2021wilds}, we use the average domain accuracy~(ADA) which is computed as follows:
\begin{eqnarray}
    \text{ADA} &= \frac{1}{|D|} \sum_{i=1}^N \text{Acc}_i \\
    \text{Acc}_i &= \frac{\text{TP}}{\text{TP} + \text{FP} + \text{FN}}
    \label{eq:ADA}
\end{eqnarray}
where $\text{Acc}_i$ denote the average accuracy corresponding to the $i^{th}$ domain and it is computed by using the confusion matrix parameters (True Positives (TP), False Positives (FP), False Negatives (FN)). Following the WiLDS benchmark~\cite{koh2021wilds} settings for GWHD 2021 dataset, we compute the confusion matrix parameters by using the Intersection of Union (IoU) threshold of 0.5 and a confidence threshold of 0.5.   

\subsection{Results}
\label{sec:results}
\begin{figure}[!ht]
    \centering
    \includegraphics[scale=0.29]{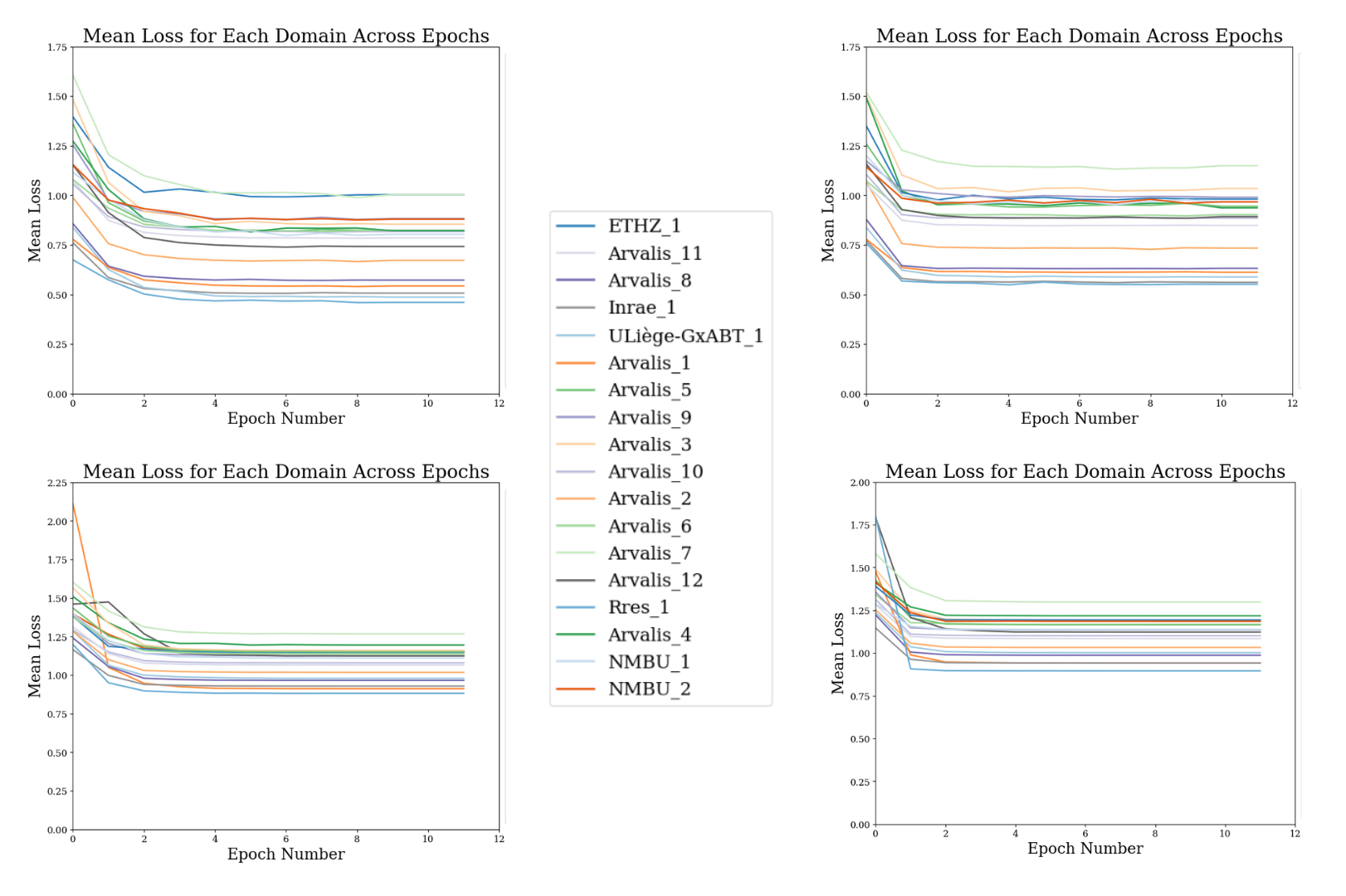}
    \caption{Training loss curves: First row: FasterRCNN, Second row: FCOS. First Column: ERM. Second Column: DP (ours).}
    \label{fig:train_loss_curves}
\end{figure}

We closely follow the experiment protocols as described in~\cite{koh2021wilds} to prove the improved generalisation performance of the proposed approach. Even though there are several state-of-the-art approaches, we primarily compare against the techniques whose contribution is primarily related to loss function as our contribution is related to the loss function. By using the hyperparameters as mentioned in the WiLDS benchmark, we attempted to replicate the experiments for the FasterRCNN detector and also train the FCOS detector. Fig~\ref{fig:train_loss_curves} shows the average training loss curves corresponding to the FasterRCNN and FCOS detectors for each of the individual source domains in the first and second rows, respectively. The left and right sides of the legend of Fig.~\ref{fig:train_loss_curves} corresponds to the loss curves for ERM and DP, respectively. It is to be noted that the training loss curves for some of the domains do not exhibit a monotonously decreasing pattern when ERM is used. In contrast, the proposed approach shows a monotonously decreasing pattern in the majority of the source domains. This pattern of training loss curves indicate is primarily due to the adaptive weighting mechanism in our approach, which forces the detector to perform well across all source domains simultaneously, thereby helping it to learn better domain-invariant features.
 
%It is to be noted that the mean loss for each of the domains is higher for the proposed approach over the ERM and this effect can be attributed to the fact that our approach is attempting to avoid overfitting by making sure that the loss is minimised across domains. (Include train-val curves to demonstrate the overfitting prevention).  

\begin{comment}
\begin{figure}[!ht]
\centering
    \begin{tabular}{cc}
    \includegraphics[scale=0.15]{media/FRCNN_baseline_train_loss.jpeg} & 
    \includegraphics[scale=0.15]{media/FRCNN_DP_train_loss.jpeg}   \\
    \includegraphics[scale=0.2]{media/FCOS_WITHOUT_DP.png} & 
    \includegraphics[scale=0.2]{media/FCOS_WITH_DP.png} \\
     ERM  & DP (Ours) \\
    \end{tabular}
    \caption{Training loss curves: First row: FasterRCNN, Second row: FCOS.}
    \label{fig:train_loss_curves}
\end{figure}    
\end{comment}

\begin{figure}[!ht]
    \centering
    \begin{tabular}{cc}
     \includegraphics[scale=0.3]{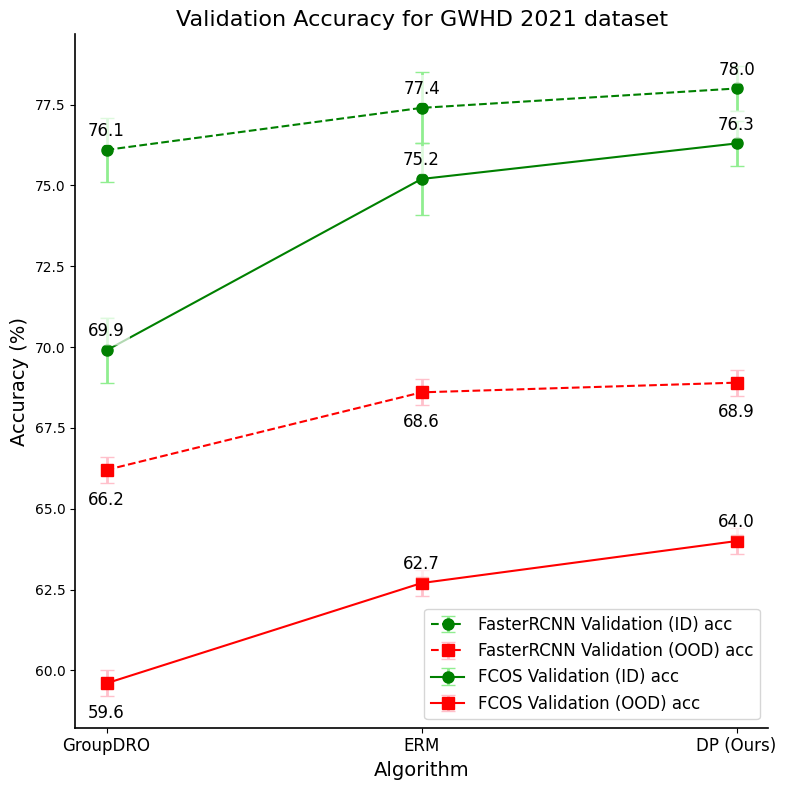}  &  
     \includegraphics[scale=0.3]{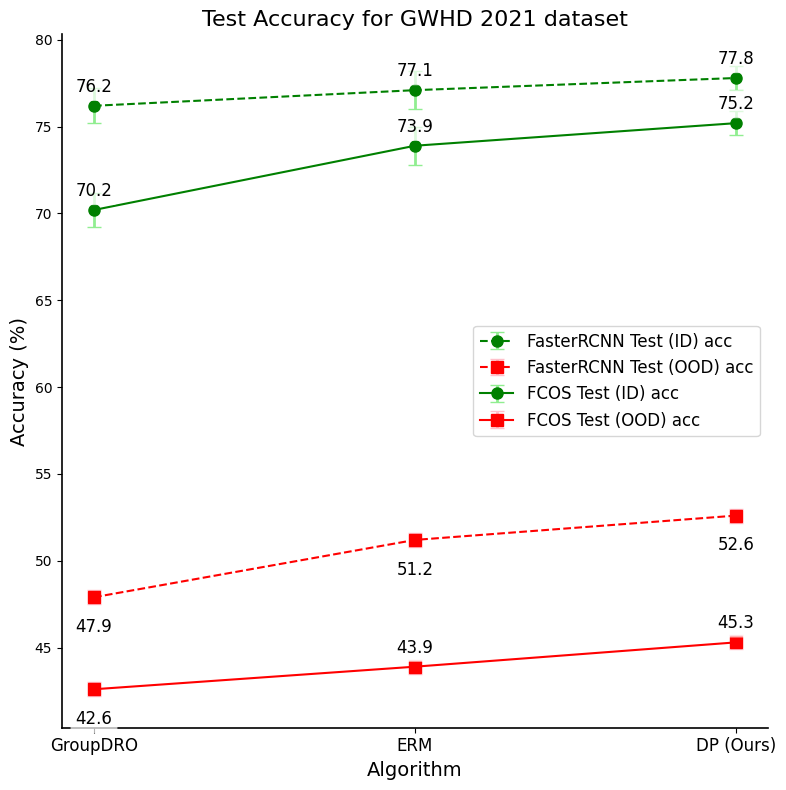}   \\
      (a)   &  (b)
    \end{tabular}
    \caption{Quantitative Analysis for GWHD 2021 dataset for FasterRCNN detector. (a) Validation set. (b) Test set.}
    \label{fig:quantitative_analysis}
\end{figure}

Fig.~\ref{fig:quantitative_analysis} represents the quantitative analysis for the performance of the proposed approach on the official in-distribution (ID) and the out-of-distribution (OOD) splits of GWHD 2021 dataset. The experimental results demonstrate that the proposed Domain Penalization (DP) method achieves higher accuracy compared to ERM and GroupDRO in all the experimental settings. Specifically, DP achieved a test OOD accuracy of 52.6$\%$, surpassing ERM's 51.2$\%$ and GroupDRO's 47.9$\%$ (see Fig.~\ref{fig:quantitative_analysis}). While for FCOS, we surpass the best performing ERM by 1.4$\%$ on the test OOD setting. This improvement is crucial for real-world applications where the distribution of test data often deviates from that of the training data. The superior OOD performance of DP suggests that it is more effective in generalising to new, unseen environments, making it a valuable approach for scenarios requiring robust domain generalisation. While GroupDRO also aims to improve robustness by focusing on the worst-case group performance, our experimental results indicate that DP's dynamic penalisation based on average domain loss is more effective. The higher accuracy achieved by DP can be attributed due to its adaptive weighting mechanism, which more effectively balances the training across multiple domains and thus achieving improved generalisation.

\begin{comment}
\begin{table}[!ht]
    \centering
    \begin{tabular}{|c|c|c|c|c|}
    \hline
     Algorithm & Validation (ID) acc & Validation (OOD) acc & Test (ID) acc & Test (OOD) acc \\
    \hline
     ERM & 77.4 (1.1) & 68.6 (0.4) & 77.1 (0.5) & 51.2 (1.8) \\
    \hline
     GroupDRO & 76.1 (1.0) & 66.2 (0.4) & 76.2 (0.8) & 47.9 (2.0) \\
    \hline
     DP (Ours) & \textbf{78.0 (0.7)} & \textbf{68.9 (0.4)} & \textbf{77.8 (0.2)}  & \textbf{52.6 (0.5)} \\
    \hline
    \end{tabular}
    \caption{Quantitative Analysis for GWHD 2021 dataset for FasterRCNN detector.}
    \label{tab:gwhd_ood_id}
\end{table}   
\end{comment}

\begin{table}
    \centering
    \begin{tabular}{|c|c|c|}
    \hline
    Setting  &  Algorithm & Test Accuracy ($\%$) \\
    \hline
    Official   & ERM & 49.6 (1.9) \\
    \hline
    Official   & DP (Ours) & \textbf{52.7 (1.5)} \\
    \hline
    Mixed-to-test  & ERM & 63.3 (1.7) \\
    \hline
    Mixed-to-test  & DP (Ours) & \textbf{66.7 (1.6)} \\
    \hline
    \end{tabular}
    \caption{Mixed to test comparison for GWHD 2021 dataset. All the evaluations reported in this table are performed on mixed-test split. The model is trained on official train split for the first two rows, while the model is trained on mixed-train set for the last two rows.}
    \label{tab:mixed_to_test_overall}
\end{table}

\begin{table}[!h]
    \centering
    \begin{tabular}{|c|c|c|c|c|c|c|c|c|}
    \hline
    Sess ID  & Country & Images & ID acc & ID acc & OOD acc & OOD acc & Ggap & Ggap \\
      &  & & (ERM) & (DP) & (ERM) & (DP) & (ERM) & (DP) \\
    \hline
     43 & Mexico & 35 & 63.1 (1.4) & 65.4 (1.2) & 48.0 (2.6)& 52.0 (2.5)  & 15.1 & \textbf{13.4}  \\    
    \hline
     44 & Mexico & 39 & 76.1 (0.9) & 77.9 (0.9) & 58.2 (3.6)& 61.1 (3.6) & 17.9 & \textbf{16.8} \\    
    \hline
     45 & Mexico & 30 & 65.6 (3.1) & 67.7 (3.2) & 63.3 (2.1)& 66.0 (2.2) & 2.3 & \textbf{1.7}   \\    
    \hline
     39 & US & 50 & 73.5 (1.1) & 74.8 (1.0) & 53.2 (2.3) & 55.1 (2.2) & 20.3  & \textbf{19.7} \\    
    \hline
     40 & US & 50 & 73.6 (0.5) & 76.2 (0.3) & 52.7 (2.7) & 55.8 (2.5) & 20.9 & \textbf{20.4}  \\    
    \hline
     41 & US & 48 & 73.3 (1.2) & 74.2 (1.1) & 48.9 (3.0) & 52.5 (3.1) & 24.4 & \textbf{21.7} \\    
    \hline
     42 & US & 30 & 68.3 (0.6) & 70.9 (0.6) & 48.7 (3.5) & 52.0 (3.1) & 19.6 & \textbf{18.9} \\    
    \hline
     37 & US & 72 & 48.9 (0.5) & 51.6 (0.6) & 17.9 (4.3) & 23.3 (4.3) & 31.0 & \textbf{28.3} \\    
    \hline
     38 & US & 53 & 34.7 (1.3) & 42.3 (1.5) & 16.0 (3.4) & 18.3 (3.6) & \textbf{18.7} & 24  \\    
    \hline
     46   & Canada & 100 & 78.3 (0.8)& 80.0 (0.7) & 77.1 (1.4) & 73.8 (2.3) & \textbf{1.2} & 6.2 \\    
    \hline
     26      & Australia & 11 & 41.8 (1.4) & 42.2 (1.5) & 29.0 (1.0) & 31.4 (0.9)  & 12.8 & \textbf{10.8}  \\    
    \hline
     27      & Australia & 8 & 81.6 (12.5) & 93.8 (10.9) & 76.5 (14.4) & 80.2 (12.2)  & \textbf{5.1} & 13.6  \\    
    \hline
     28      & Australia & 7 & 56.4 (13.8) & 68.6 (14.1) & 54.3 (10.0) & 65.9 (9.0)  & \textbf{2.1} & 2.7  \\    
    \hline
     29      & Australia & 15 & 68.8 (0.5) & 70.8 (0.3) & 60.6 (1.3) & 62.5 (1.2) & \textbf{8.2} & 8.3  \\    
    \hline
     30      & Australia & 15 & 54.4 (2.1) & 61.1 (1.9)  & 38.6 (2.1) & 38.5 (2.5) & \textbf{15.8} & 38.1  \\    
    \hline
     31      & Australia & 15 & 75.8 (1.1) & 76.2 (1.1) & 71.9 (0.7) & 72.4 (0.7) & 3.9 & \textbf{3.8}  \\    
    \hline
     32      & Australia & 9 & 68.9 (0.6) & 70.7 (0.7) & 62.8 (2.5) & 66.2 (2.3)  & 6.1 & \textbf{4.5}\\    
    \hline
     33      & Australia & 21 & 58.6 (0.6) & 60.9 (0.5) & 46.5 (2.1) & 50.6 (2.2) & 11.1 & \textbf{10.3} \\    
    \hline
     34      & Australia & 17 & 54.7 (1.5) & 56.0 (1.3)  & 43.6 (2.1) & 47.5 (2.0)  & 11.1 & \textbf{8.5} \\    
    \hline
     35      & Australia & 53 & 61.7 (0.8) & 64.8 (0.7) & 39.6 (2.5) & 44.0 (2.1) & 22.1 & \textbf{20.8} \\    
    \hline
     36      & Australia & 42 & 50.4 (1.5) & 54.0 (1.3) & 33.5 (2.7)& 37.7 (2.8)  & 16.9 & \textbf{16.3}   \\    
    \hline
     Total      &  & 720 & 63.3 (1.7) & 66.7 (1.6) & 49.6 (1.9) & 52.7 (1.5) & \textbf{13.7} & 14.0 \\    
    \hline
    
    \end{tabular}
    \caption{Performance analysis on each of the individual domains of mixed-to-test setting for FasterRCNN detector trained on GWHD 2021 dataset. This is a detailed version of Table.~\ref{tab:mixed_to_test_overall}. Here Ggap denotes the generalisation gap between the ID and OOD performance.}
    \label{tab:mixed_to_test_detailed}
\end{table}
Even though our proposed approach has shown improved in-distribution (ID) and out-of-distribution (OOD) performance, the performance evaluations were conducted on different test splits. To accurately measure the generalisation gap, evaluation should be performed on the same split while training two different models in ID and OOD settings, respectively. Following the experimental protocols established in the WiLDS benchmark~\cite{koh2021wilds}, we report our experimental analysis for measuring the generalisation gap in Tables~\ref{tab:mixed_to_test_overall} and \ref{tab:mixed_to_test_detailed}.  Table~\ref{tab:mixed_to_test_overall} summarise the overall test accuracy on the mixed-test split. The first two rows represent the test accuracy of models trained only on the official train split of the GWHD 2021 dataset using ERM and our proposed approach, respectively. The last two rows show the test accuracy of models trained on a mix of the official train and test splits of the GWHD 2021 dataset. The results clearly indicate that our proposed DP approach outperforms ERM in both scenarios. Furthermore, Table~\ref{tab:mixed_to_test_detailed} offers a comprehensive accuracy analysis for each test domain. The proposed DP approach not only achieves better average accuracy than ERM but also performs well across majority of the individual  test domains in both the ID and OOD settings. This highlights the effectiveness of the dynamic weighting scheme employed in DP. Interestingly, DP reduces the ID-OOD gap compared to ERM for 15 out of 21 test domains. This reduction in the gap suggests that the features learned by the detector network using our proposed loss function are more stable, consistent, and transferable to unseen domains.

\section{Conclusions}
\label{sec:conclusions}
In this paper, we introduce a domain penalization framework aimed at enhancing the out-of-distribution generalization performance of object detection networks within a multi-source domain context. We assign weights to each domain, with values computed based on the object detector’s performance on the corresponding domain. Our approach demonstrates superior performance compared to other loss function-based techniques used in the WiLDS benchmark. We demonstrate that the proposed method not only improves accuracy on test domains but also narrows the ID-OOD gap for most test domains. This indicates that our approach is more stable and predictable on new, unseen domains during the training process. Although our experiments are limited to the object detection scenario, the same framework can be extended to other datasets in the WiLDS benchmark, which are primarily focused on image classification. This extension will be investigated in future research. 

%\subsubsection{Acknowledgements} Please place your acknowledgments at
%the end of the paper, preceded by an unnumbered run-in heading (i.e.
%3rd-level heading).

\bibliographystyle{splncs04}
\bibliography{mybibliography}

\begin{thebibliography}{10}
\providecommand{\url}[1]{\texttt{#1}}
\providecommand{\urlprefix}{URL }
\providecommand{\doi}[1]{https://doi.org/#1}

\bibitem{arjovsky2019invariant}
Arjovsky, M., Bottou, L., Gulrajani, I., Lopez-Paz, D.: Invariant risk minimization. arXiv preprint arXiv:1907.02893  (2019)

\bibitem{beery2018recognition}
Beery, S., Van~Horn, G., Perona, P.: Recognition in terra incognita. In: Proceedings of the European conference on computer vision (ECCV). pp. 456--473 (2018)

\bibitem{carion2020end}
Carion, N., Massa, F., Synnaeve, G., Usunier, N., Kirillov, A., Zagoruyko, S.: End-to-end object detection with transformers. In: European conference on computer vision. pp. 213--229. Springer (2020)

\bibitem{chen2018domain}
Chen, Y., Li, W., Sakaridis, C., Dai, D., Van~Gool, L.: Domain adaptive faster r-cnn for object detection in the wild. In: Proceedings of the IEEE conference on computer vision and pattern recognition. pp. 3339--3348 (2018)

\bibitem{david2021global}
David, E., Serouart, M., Smith, D., Madec, S., Velumani, K., Liu, S., Wang, X., Espinosa, F.P., Shafiee, S., Tahir, I.S., et~al.: Global wheat head dataset 2021: more diversity to improve the benchmarking of wheat head localization methods. arXiv preprint arXiv:2105.07660  (2021)

\bibitem{deng2009imagenet}
Deng, J., Dong, W., Socher, R., Li, L.J., Li, K., Fei-Fei, L.: Imagenet: A large-scale hierarchical image database. In: 2009 IEEE conference on computer vision and pattern recognition. pp. 248--255. Ieee (2009)

\bibitem{faster2015towards}
Faster, R.: Towards real-time object detection with region proposal networks. Advances in neural information processing systems  \textbf{9199}(10.5555),  2969239--2969250 (2015)

\bibitem{ganin2016domain}
Ganin, Y., Ustinova, E., Ajakan, H., Germain, P., Larochelle, H., Laviolette, F., March, M., Lempitsky, V.: Domain-adversarial training of neural networks. Journal of machine learning research  \textbf{17}(59),  1--35 (2016)

\bibitem{hindel2023inod}
Hindel, J., Gosala, N., Bregler, K., Valada, A.: Inod: Injected noise discriminator for self-supervised representation learning in agricultural fields. IEEE Robotics and Automation Letters  (2023)

\bibitem{hu2018does}
Hu, W., Niu, G., Sato, I., Sugiyama, M.: Does distributionally robust supervised learning give robust classifiers? In: International Conference on Machine Learning. pp. 2029--2037. PMLR (2018)

\bibitem{ilteralp2021deep}
Ilteralp, M., Ariman, S., Aptoula, E.: A deep multitask semisupervised learning approach for chlorophyll-a retrieval from remote sensing images. Remote Sensing  \textbf{14}(1), ~18 (2021)

\bibitem{jing2020self}
Jing, L., Tian, Y.: Self-supervised visual feature learning with deep neural networks: A survey. IEEE transactions on pattern analysis and machine intelligence  \textbf{43}(11),  4037--4058 (2020)

\bibitem{karunanayake2024out}
Karunanayake, N., Gunawardena, R., Seneviratne, S., Chawla, S.: Out-of-distribution data: An acquaintance of adversarial examples--a survey. arXiv preprint arXiv:2404.05219  (2024)

\bibitem{khoee2024domain}
Khoee, A.G., Yu, Y., Feldt, R.: Domain generalization through meta-learning: A survey. arXiv preprint arXiv:2404.02785  (2024)

\bibitem{koh2021wilds}
Koh, P.W., Sagawa, S., Marklund, H., Xie, S.M., Zhang, M., Balsubramani, A., Hu, W., Yasunaga, M., Phillips, R.L., Gao, I., et~al.: Wilds: A benchmark of in-the-wild distribution shifts. In: International conference on machine learning. pp. 5637--5664. PMLR (2021)

\bibitem{li2022learning}
Li, W.H., Liu, X., Bilen, H.: Learning multiple dense prediction tasks from partially annotated data. In: Proceedings of the IEEE/CVF Conference on Computer Vision and Pattern Recognition. pp. 18879--18889 (2022)

\bibitem{li2018deep}
Li, Y., Tian, X., Gong, M., Liu, Y., Liu, T., Zhang, K., Tao, D.: Deep domain generalization via conditional invariant adversarial networks. In: Proceedings of the European conference on computer vision (ECCV). pp. 624--639 (2018)

\bibitem{lin2021domain}
Lin, C., Yuan, Z., Zhao, S., Sun, P., Wang, C., Cai, J.: Domain-invariant disentangled network for generalizable object detection. In: Proceedings of the IEEE/CVF international conference on computer vision. pp. 8771--8780 (2021)

\bibitem{lin2017focal}
Lin, T.Y., Goyal, P., Girshick, R., He, K., Doll{\'a}r, P.: Focal loss for dense object detection. In: Proceedings of the IEEE international conference on computer vision. pp. 2980--2988 (2017)

\bibitem{lin2014microsoft}
Lin, T.Y., Maire, M., Belongie, S., Hays, J., Perona, P., Ramanan, D., Doll{\'a}r, P., Zitnick, C.L.: Microsoft coco: Common objects in context. In: Computer Vision--ECCV 2014: 13th European Conference, Zurich, Switzerland, September 6-12, 2014, Proceedings, Part V 13. pp. 740--755. Springer (2014)

\bibitem{liu2020towards}
Liu, H., Song, P., Ding, R.: Towards domain generalization in underwater object detection. In: 2020 IEEE international conference on image processing (ICIP). pp. 1971--1975. IEEE (2020)

\bibitem{liu2021towards}
Liu, J., Shen, Z., He, Y., Zhang, X., Xu, R., Yu, H., Cui, P.: Towards out-of-distribution generalization: A survey. arXiv preprint arXiv:2108.13624  (2021)

\bibitem{long2015learning}
Long, M., Cao, Y., Wang, J., Jordan, M.: Learning transferable features with deep adaptation networks. In: International conference on machine learning. pp. 97--105. PMLR (2015)

\bibitem{nguyen2024tackling}
Nguyen, V.D., Mirza, S., Zakeri, A., Gupta, A., Khaldi, K., Aloui, R., Mantini, P., Shah, S.K., Merchant, F.: Tackling domain shifts in person re-identification: A survey and analysis. In: Proceedings of the IEEE/CVF Conference on Computer Vision and Pattern Recognition. pp. 4149--4159 (2024)

\bibitem{niu2024survey}
Niu, Z., Ouyang, S., Xie, S., Chen, Y.w., Lin, L.: A survey on domain generalization for medical image analysis. arXiv preprint arXiv:2402.05035  (2024)

\bibitem{noroozi2016unsupervised}
Noroozi, M., Favaro, P.: Unsupervised learning of visual representations by solving jigsaw puzzles. In: European conference on computer vision. pp. 69--84. Springer (2016)

\bibitem{pan2009survey}
Pan, S.J., Yang, Q.: A survey on transfer learning. IEEE Transactions on knowledge and data engineering  \textbf{22}(10),  1345--1359 (2009)

\bibitem{redmon2016you}
Redmon, J., Divvala, S., Girshick, R., Farhadi, A.: You only look once: Unified, real-time object detection. In: Proceedings of the IEEE conference on computer vision and pattern recognition. pp. 779--788 (2016)

\bibitem{saenko2010adapting}
Saenko, K., Kulis, B., Fritz, M., Darrell, T.: Adapting visual category models to new domains. In: Computer Vision--ECCV 2010: 11th European Conference on Computer Vision, Heraklion, Crete, Greece, September 5-11, 2010, Proceedings, Part IV 11. pp. 213--226. Springer (2010)

\bibitem{sagawa2019distributionally}
Sagawa, S., Koh, P.W., Hashimoto, T.B., Liang, P.: Distributionally robust neural networks for group shifts: On the importance of regularization for worst-case generalization. arXiv preprint arXiv:1911.08731  (2019)

\bibitem{seemakurthy2023domain_fcos}
Seemakurthy, K., Bosilj, P., Aptoula, E., Fox, C.: Domain generalised fully convolutional one stage detection. In: 2023 IEEE International Conference on Robotics and Automation (ICRA). pp. 7002--7009. IEEE (2023)

\bibitem{seemakurthy2023domain_frcnn}
Seemakurthy, K., Fox, C., Aptoula, E., Bosilj, P.: Domain generalised faster r-cnn. In: Proceedings of the AAAI Conference on Artificial Intelligence. vol.~37, pp. 2180--2190 (2023)

\bibitem{sun2016deep}
Sun, B., Saenko, K.: Deep coral: Correlation alignment for deep domain adaptation. In: Computer Vision--ECCV 2016 Workshops: Amsterdam, The Netherlands, October 8-10 and 15-16, 2016, Proceedings, Part III 14. pp. 443--450. Springer (2016)

\bibitem{tobin2017domain}
Tobin, J., Fong, R., Ray, A., Schneider, J., Zaremba, W., Abbeel, P.: Domain randomization for transferring deep neural networks from simulation to the real world. In: 2017 IEEE/RSJ international conference on intelligent robots and systems (IROS). pp. 23--30. IEEE (2017)

\bibitem{tzeng2014deep}
Tzeng, E., Hoffman, J., Zhang, N., Saenko, K., Darrell, T.: Deep domain confusion: Maximizing for domain invariance. arXiv preprint arXiv:1412.3474  (2014)

\bibitem{ullah2023ssmd}
Ullah, Z., Usman, M., Latif, S., Khan, A., Gwak, J.: Ssmd-unet: semi-supervised multi-task decoders network for diabetic retinopathy segmentation. Scientific Reports  \textbf{13}(1), ~9087 (2023)

\bibitem{wang2022generalizing}
Wang, J., Lan, C., Liu, C., Ouyang, Y., Qin, T., Lu, W., Chen, Y., Zeng, W., Philip, S.Y.: Generalizing to unseen domains: A survey on domain generalization. IEEE transactions on knowledge and data engineering  \textbf{35}(8),  8052--8072 (2022)

\bibitem{yue2019domain}
Yue, X., Zhang, Y., Zhao, S., Sangiovanni-Vincentelli, A., Keutzer, K., Gong, B.: Domain randomization and pyramid consistency: Simulation-to-real generalization without accessing target domain data. In: Proceedings of the IEEE/CVF international conference on computer vision. pp. 2100--2110 (2019)

\bibitem{zech2018variable}
Zech, J.R., Badgeley, M.A., Liu, M., Costa, A.B., Titano, J.J., Oermann, E.K.: Variable generalization performance of a deep learning model to detect pneumonia in chest radiographs: a cross-sectional study. PLoS medicine  \textbf{15}(11),  e1002683 (2018)

\bibitem{zhou2022domain}
Zhou, K., Liu, Z., Qiao, Y., Xiang, T., Loy, C.C.: Domain generalization: A survey. IEEE Transactions on Pattern Analysis and Machine Intelligence  \textbf{45}(4),  4396--4415 (2022)

\end{thebibliography}
\end{document}